\documentclass[runningheads]{llncs}
\usepackage[T1]{fontenc}
\usepackage{graphicx,verbatim}
\usepackage{amsmath}
\usepackage{amssymb}
\usepackage{svg}
\usepackage{float}
\usepackage{marvosym}

\usepackage{booktabs}
\usepackage{multirow}
\usepackage{siunitx}
\usepackage{xcolor}
\usepackage{capt-of}
\usepackage{amsfonts}
\usepackage{bm}
\usepackage{xspace}
\makeatletter
\DeclareRobustCommand\onedot{\futurelet\@let@token\@onedot}
\def\@onedot{\ifx\@let@token.\else.\null\fi\xspace}
\def\eg{\emph{e.g}\onedot}

\makeatother

\usepackage{pifont}

\newcommand{\best}[1]{\textbf{#1}}
\newcommand{\second}[1]{\underline{#1}}

\usepackage[hidelinks]{hyperref}
\begin{document}
\title{Steer the Sampling, Not the Kernel Grid: Geometry-Guided Sampling Operator for Volumetric Segmentation}
%\titlerunning{Abbreviated paper title}
% If the paper title is too long for the running head, you can set
% an abbreviated paper title here
%
\begin{comment}  %% Removed for anonymized MICCAI submission
\author{First Author\inst{1}\orcidID{0000-1111-2222-3333} \and
Second Author\inst{2,3}\orcidID{1111-2222-3333-4444} \and
Third Author\inst{3}\orcidID{2222--3333-4444-5555}}
%
\authorrunning{F. Author et al.}
% First names are abbreviated in the running head.
% If there are more than two authors, 'et al.' is used.
%
\institute{Princeton University, Princeton NJ 08544, USA \and
Springer Heidelberg, Tiergartenstr. 17, 69121 Heidelberg, Germany
\email{lncs@springer.com}\\
\url{http://www.springer.com/gp/computer-science/lncs} \and
ABC Institute, Rupert-Karls-University Heidelberg, Heidelberg, Germany\\
\email{\{abc,lncs\}@uni-heidelberg.de}}

\end{comment}

\author{ Sizhe Wang\inst{1}\textsuperscript{(\Letter)} \and Himashi Peiris\inst{1} \and Zhaolin Chen\inst{1}\textsuperscript{(\Letter)}} 
%index{Wang, Sizhe} 
%index{Peiris, Himashi} 
%index{Chen, Zhaolin} 

\authorrunning{S. Wang et al.} 

\institute{ Department of Data Science \& AI, Faculty of Information Technology, Monash University, Clayton, VIC, Australia\\ \email{\{sizhe.wang,zhaolin.chen\}@monash.edu}
}
  
\maketitle              % typeset the header of the contribution
\begin{abstract}

Accurate 3D segmentation is central to quantitative lesion assessment and anatomy mapping for clinical planning and follow-up.
Thin, elongated, and fine anatomical/pathological structures (\eg, vessels) are a particularly stress case: a one-voxel boundary error can disconnect a branch and change clinically relevant topology.
In encoder--decoder networks (\eg, U-Net), repeated downsampling and fixed-grid convolution blur/alias fine structures and weaken orientation cues, so early mistakes propagate across scales.
We propose a geometry-guided local operator that \emph{steers where features are sampled} (rather than deforming convolutional kernels) under a single formulation for both feature refinement (stride $1$) and resolution reduction (stride $>1$).
At each voxel, it predicts a local orientation and bounded step sizes, samples symmetrically along these directions, and transforms paired samples into compact geometric/boundary cues with lightweight mixing; a cross-scale consensus aligns encoder and decoder features at skip connections to reduce geometric mismatch.
Replacing all stride $1$ and stride $2$ operators in a 3D U-Net yields consistent improvements on BraTS, MSD Hepatic Vessel, and TDSC-ABUS, with notably better boundary metrics (\eg, BraTS Dice $86.1\!\rightarrow\!88.9$, HD95 $7.1\!\rightarrow\!6.2$; TDSC-ABUS HD95 $39.1\!\rightarrow\!27.8$) while largely reducing parameters ($2.3$M$\!\rightarrow\!0.8$M).
We further demonstrate that the operator plugs into other backbones (\eg nnU-Net, Swin-UNETR, and MedNeXt) without changing their macro-architectures while providing consistent performance gains.
Code available at \href{https://github.com/AndyWanng/GeoSample}{GeoSample repo}.

% Authors must provide keywords and are not allowed to remove this Keyword section.

\end{abstract}

\section{Introduction}
Accurate volumetric segmentation is a prerequisite for many downstream clinical and scientific workflows, including lesion quantification, preoperative planning, and longitudinal follow-up \cite{pham2000current,hesamian2019deep}. 
Encoder--decoder architectures in the U-Net family remain dominant because they combine multi-scale context aggregation with fine localisation through skip connections \cite{ronneberger2015u,cciccek20163d}; nnU-Net further demonstrates strong performance across diverse biomedical tasks using largely standardised backbones and robust training recipes \cite{isensee2021nnu}. 
Despite these successes, thin, elongated, or anisotropic structures remain fragile, and more broadly any task requiring precise boundary localisation can suffer when resolution changes distort local geometry~\cite{sinclair2025perivascular}.
% Such targets often occupy only a few voxels and encode clinically relevant topological features (\eg, connectivity and endpoints), and small boundary shifts can break continuity and bias quantitative measurements \cite{hu2019topology}.
Such targets often occupy only a few voxels, so small boundary shifts can break continuity and bias quantitative measurements \cite{hu2019topology}.
\emph{Therefore, rather than redesigning the backbone, we fundamentally rethink the convolution/pooling primitives that sample, aggregate, and downsample features.}

A key issue is that existing stride-based refinement and downsampling indiscriminately aggregate voxel information along fixed grids, suppressing or aliasing high-frequency details and distorting fine structures \cite{azulay2019deep,ribeiro2021convolutional}, an issue exacerbated by extreme foreground sparsity and limited sampling support. 
Though anti-aliasing strategies partially address instability under downsampling \cite{zhang2019making}, they remain largely orientation-agnostic and fail to provide a unified approach that preserves boundary geometry and directional cues across scales.

Prior work mitigates cross-scale degradation from several directions. 
Transformer based and hybrid volumetric segmenters strengthen global context modelling \cite{hatamizadeh2022unetr,hatamizadeh2021swin,liu2021swin,zhou2023nnformer,xie2021cotr,peiris2022robust}, while other studies target cross-scale calibration and alignment within U-Net pipelines \cite{yang2025dynamic}. However, these approaches operate on already-sampled features and do not reconsider the geometric bias introduced by fixed-grid refinement and downsampling.

Adaptive sampling operators move beyond fixed-grid convolutional aggregation by learning sampling offsets, \eg, deformable convolutions \cite{dai2017deformable,zhu2019deformable} and volumetric variants \cite{lee2023deformux}, as well as sparse large-kernel sampling designs such as OBELISK \cite{heinrich2019obelisk}. However, these methods predict unconstrained offsets per kernel location, lacking structured geometric regularisation and symmetry; neighbouring voxels may yield irregular sampling patterns, and the representation is not orientation-consistent.

Differentiable warping modules and equivariant or steerable architectures incorporate geometric transformations or symmetry constaints~\cite{jaderberg2015spatial,frangi1998multiscale,cohen2016group,weiler20183d,fuchs2020se}. 
However, these existing approaches are either anchored in fixed-grid convolutional backbones or impose global equivariance without explicitly modelling boundary-aligned sampling for both refinement and downsampling. In contrast, we argue sampling directions should be locally structured, rotation-consistent, and symmetrically paired, with bounded step sizes to ensure stability across scales.
% In spirit, our design is related to geometric deep learning in that it learns a voxel-wise $SO(3)$ local frame to steer symmetric sampling and extract geometric cues, while not aiming for strict rotation equivariance.

We propose \emph{GeoSample}, a \emph{geometry-guided} local operator as a unified, plug-in module for volumetric encoder--decoder networks.
Rather than deforming kernels, GeoSample predicts a voxel-wise local 3D rotation frame (an element of $\mathrm{SO}(3)$) with adaptive step sizes to steer structured symmetric sampling. The resulting average/difference signals are converted into step-size normalised differential cues that compactly summarise gradient- and curvature-like information. To reduce geometric mismatch between decoder and encoder features, we further employ a rotation-consistent cross-scale consensus mechanism.

Our contributions are threefold:
\textbf{(i)} a novel, geometry-guided local \emph{operator} (\emph{GeoSample}) that replaces fixed-grid convolution/pooling and unifies stride $1$ refinement and stride $>1$ downsampling under a single formulation;
\textbf{(ii)} a rotation-constrained local frame with adaptive step sizes that enables \emph{structured symmetric sampling} and extracts compact geometry-aware signals, improving segmentation fidelity for thin or anisotropic targets; and
\textbf{(iii)} a rotation-consistent \emph{Consensus Field} module that aligns geometry at the skip connection, improving robustness across backbones.
Fig.~\ref{fig:method_overview} provides an overview; we next detail the formulation in Sec.~\ref{sec:methodology}.

\section{Methodology}
\label{sec:methodology}

% \subsection{Overview}
% We propose a \emph{frame-conditioned differential transport} operator for volumetric segmentation.
% The operator is used in two modes:
% (i) a \emph{stride-1} mode that replaces standard convolutional blocks to maintain a geometry-aware representation without changing resolution, and
% (ii) a \emph{downsampling} mode for $s\neq(1,1,1)$ that replaces stride-based reduction while explicitly preserving directional information under aggregation.
% Both modes share the same geometric field prediction and tokenisation pipeline, which simplifies multi-scale modelling and cross-scale alignment.

\subsection{Overview and integration}
\begin{figure}[t]
  \centering
  \includegraphics[width=0.95\textwidth]{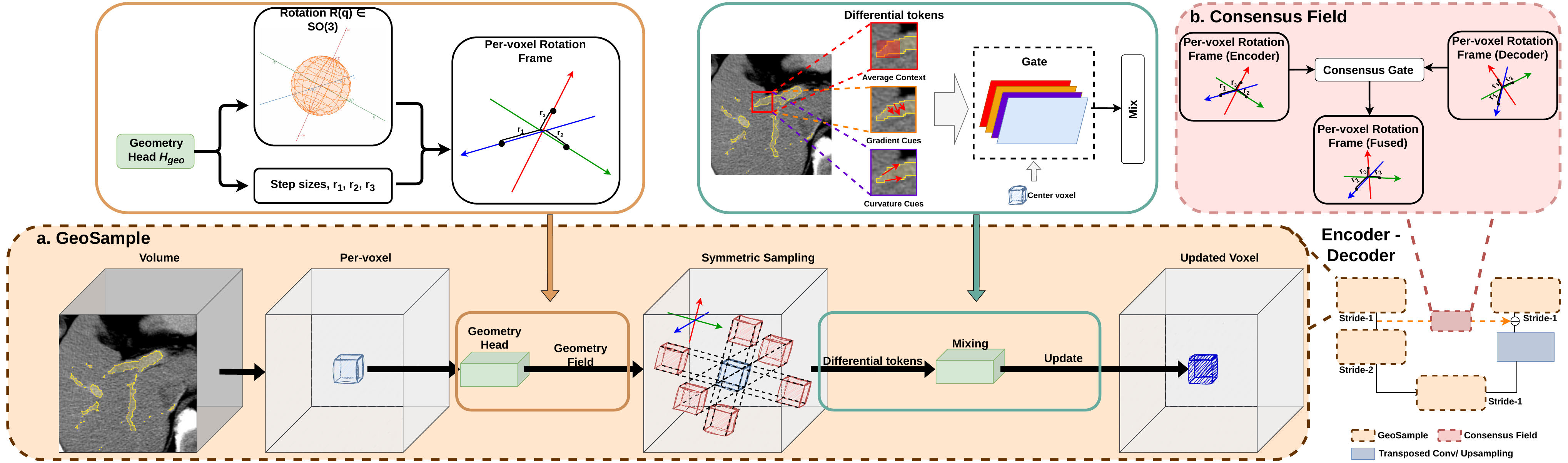}
  \caption{Overall workflow of the proposed modules. (a) \emph{GeoSample}: per-voxel geometry field, symmetric sampling, and differential-token mixing. (b) \emph{Consensus Field}: rotation-consistent fusion of geometry fields from the encoder skip and upsampled decoder features for pre-skip alignment. Shown on a U-Net, \emph{GeoSample} replaces stride-$1$ blocks and stride-$2$ downsampling; \emph{Consensus Field} is applied at skip connection.}
  \label{fig:method_overview}
\end{figure}

Given an input feature volume $X\in\mathbb{R}^{C\times D\times H\times W}$, we predict a voxel-wise geometric field $\mathcal{H}(X)$ (guiding our operator $\Psi$) and use it to steer symmetric sampling with lightweight mixing.
This yields a drop-in local operator that replaces stride $1$ refinement blocks and stride $>1$ downsampling in 3D encoder--decoder segmenters (Fig.~\ref{fig:method_overview}).

\paragraph{\textbf{Stride $1$ refinement (replacement of conv blocks).}}
For feature refinement without changing spatial resolution, we generate a geometry-conditioned residual update:
\begin{equation}
  Y = X + \Psi_{1}\!\big(X;\mathcal{H}(X)\big).
  \label{eq:stride1_use}
\end{equation}

\paragraph{\textbf{Stride $>1$ downsampling (replacement of strided conv/pooling).}}
For spatial reduction, we adopt average pooling as a stable low-pass path and augment it with a geometry-conditioned compensation term:
\begin{equation}
  Y_{\downarrow} = \operatorname{AvgPool}_{s}(Y) + \Psi_{\downarrow,s}\!\big(X;\mathcal{H}(X)\big), \qquad s\neq(1,1,1).
  \label{eq:down_use}
\end{equation}

\paragraph{\textbf{Consensus Field at skip connection.}}
At each decoder stage, we fuse the geometric field from the upsampled decoder feature with the corresponding encoder skip feature, and apply the stride $1$ operator using the fused field before concatenation.
This reduces mismatch across the skip connection (Sec.~\ref{subsec:consensus}).

% \subsection{Notation and continuous sampling}
% Let $X\in\mathbb{R}^{C\times D\times H\times W}$ be an input feature volume on a discrete voxel grid.
% We define its continuous extension by differentiable trilinear interpolation
% \begin{equation}
%   x(\mathbf{p})=\mathcal{I}(X,\mathbf{p})\in\mathbb{R}^{C}, \qquad \mathbf{p}\in\mathbb{R}^{3},
% \end{equation}
% implemented with \texttt{grid\_sample}.
% We denote average pooling with kernel size and stride $s=(s_z,s_y,s_x)$ by $\operatorname{AvgPool}_s(\cdot)$.
% A small constant $\epsilon>0$ is used for numerical stability.

\subsection{Geometry field prediction}\label{subsec:field}
All operations are applied pointwise at voxel locations; we omit the spatial index for readability.
A geometry head $H_{geo}$ predicts a rotation and step-sizes:
\begin{equation}
  \mathcal{H}(X)=\{R(X), \mathbf{r}(X)\},\  R\in SO(3),\ \mathbf{r}=[r_1,\ldots,r_K]^\top\in[r_{\min},r_{\max}]^K .
  \label{eq:theta_def}
\end{equation}
Internally, we represent $R$ by a unit quaternion $q\in\mathbb{S}^3$ (with $R=\mathcal{R}(q)$); this is used only for rotation fusion in Sec.~\ref{subsec:consensus}.
Let $\{\mathbf{e}_k\}_{k=1}^{3}$ be the canonical axes of the \emph{image cartesian} coordinate system.
We take $\mathbf{u}_k = R\,\mathbf{e}_k$ as the first $K\le 3$ unit directions ($K=3$ in all experiments, yielding a full local 3D frame) and define sampling offsets
\begin{equation}
  \boldsymbol{\delta}_k = r_k\,\mathbf{u}_k \in \mathbb{R}^3 .
  \label{eq:offset_def}
\end{equation}
Importantly, $\mathbf{u}_k$ are expressed in the same coordinate system as the voxel grid; the frame is used only to steer sampling directions (we do not reparameterise features into a new coordinate system).

\subsection{Symmetric sampling and oriented differentials}\label{subsec:stencils}
We view $X$ as a continuous field $x(\cdot)$ via differentiable trilinear interpolation.
For each direction $\mathbf{u}_k$ and step size $r_k$ (both predicted at voxel location $\mathbf{p}$), we sample symmetrically:
\begin{equation}
  x_k^{+}=x(\mathbf{p}+\boldsymbol{\delta}_k),\qquad x_k^{-}=x(\mathbf{p}-\boldsymbol{\delta}_k),
\end{equation}
where $\mathbf{p}\in\mathbb{R}^3$ is a voxel coordinate.
In the continuous domain, the first- and second-order directional differentials along $\mathbf{u}_k$ are
\begin{equation}
  \partial_{\mathbf{u}_k}x = \mathbf{u}_k^\top\nabla x,
  \qquad
  \partial_{\mathbf{u}_k}^2 x = \mathbf{u}_k^\top H_x\,\mathbf{u}_k ,
  \label{eq:pde_view}
\end{equation}
and we approximate them by symmetric finite differences with step-size normalisation:
\begin{equation}
  a_k=\tfrac{1}{2}(x_k^{+}+x_k^{-}),\qquad
  d_k=\frac{x_k^{+}-x_k^{-}}{2(r_k+\epsilon)},\qquad
  s_k=\frac{x_k^{+}-2x(\mathbf{p})+x_k^{-}}{(r_k+\epsilon)^2},
  \label{eq:diff_tokens}
\end{equation}
where $\epsilon>0$ is a small constant for numerical stability.
Here, $a_k$ is an even (context) response, $d_k$ is an odd (directional-change) response, and $s_k$ captures even (2nd-order) curvature-like change. The symmetry makes the odd/even components explicit, while step-size normalisation makes the cues comparable across scales.

\subsection{Compact differential tokens and mixing}\label{subsec:tokens}
From $\{d_k\}$ we form a gradient-like cue $G$ expressed in the \emph{image Cartesian} axes, and from $\{s_k\}$ we form a curvature/Laplacian-like cue $L$:
\begin{equation}
  G = \sum_{k=1}^{K} \mathbf{u}_k \otimes d_k \in \mathbb{R}^{3\times C},
  \qquad
  L = \tfrac{1}{K}\sum_{k=1}^{K} s_k \in \mathbb{R}^{C}.
  \label{eq:GL_def}
\end{equation}
\noindent\textbf{Sign stability.}
One key design is symmetric paired sampling, which makes the directional cue \emph{sign-invariant by construction}:
$\mathbf{u}_k\!\rightarrow\!-\mathbf{u}_k \Rightarrow d_k\!\rightarrow\!-d_k$ and
$(-\mathbf{u}_k)\otimes(-d_k)=\mathbf{u}_k\otimes d_k$.
This yields stable directional tokens without frame-sign jitter.

We denote the center sample by $x_0$ and collect tokens
\begin{equation}
\mathcal{T}=
\Big[
\underbrace{x_0}_{\text{center}},
\underbrace{a_1,\ldots,a_K}_{\text{even / context}},
\underbrace{G_x,G_y,G_z}_{\text{odd / gradient-like}},
\underbrace{L}_{\text{even / curvature-like}}
\Big].
\label{eq:T_def}
\end{equation}

We apply a lightweight token-wise gating followed by $1\times1\times1$ mixing to obtain an update $\Delta$:
\begin{equation}
  \Delta = \mathrm{Mix}\big(\mathrm{Gate}(\mathcal{T}) \odot \mathcal{T}\big),
  \qquad
  \Psi_{1}(X;\mathcal{H})=\Delta.
  \label{eq:mix_highlevel}
\end{equation}
The specific realisation of $\mathrm{Gate}(\cdot)$ and $\mathrm{Mix}(\cdot)$ is summarised in Sec.~\ref{subsec:impl}.

\subsection{Downsampling with directional-signal preservation}\label{subsec:fuse_down}
For downsampling, we first compute the stride $1$ refined feature $Y$ in \eqref{eq:stride1_use}.
Since odd directional signals can undesirably cancel under naive averaging, we deliberately preserve their magnitude by pooling $|G|$:
\begin{equation}
\bar{\mathcal{T}}_{\downarrow}=
\operatorname{AvgPool}_{s}\Big(
\big[
X,
\underbrace{a_1,\ldots,a_K}_{\text{even terms}},
\underbrace{|G_x|,|G_y|,|G_z|}_{\text{odd energy}},
\underbrace{L}_{\text{even / curvature-like}}
\big]
\Big).
\end{equation}
and predict a coarse compensation term
\begin{equation}
  \Psi_{\downarrow,s}(X;\mathcal{H})=\mathrm{Mix}_{\downarrow,s}\big(\mathrm{Gate}_{\downarrow,s}(\bar{\mathcal{T}}_{\downarrow})\odot \bar{\mathcal{T}}_{\downarrow}\big).
\end{equation}

\subsection{Rotation-consistent Consensus Field and integration}\label{subsec:consensus}
To reduce geometric mismatch between upsampled decoder features and encoder skip features, we fuse two local \emph{rotation fields}: a current field $\mathcal{H}=\{R,\mathbf{r}\}$ from the upsampled decoder features with a reference field $\mathcal{H}^{*}=\{R^{*},\mathbf{r}^{*}\}$ predicted from encoder skip features, where $R,R^{*}\in \mathrm{SO}(3)$ and $\mathbf{r},\mathbf{r}^{*}\in[r_{\min},r_{\max}]^{K}$.
For rotation fusion we use their unit-quaternion representations $q,q^{*}\in\mathbb{S}^{3}$ (with $R=\mathcal{R}(q)$ and $R^{*}=\mathcal{R}(q^{*})$).

We compute a consensus gate $\omega\in(0,1)$ based on step-size statistics and quaternion similarity:
\begin{equation}
  c = |\langle q, q^{*}\rangle|,\
  \bar r = \tfrac{1}{K}\sum_{k=1}^{K} r_k,\
  \bar r^{*} = \tfrac{1}{K}\sum_{k=1}^{K} r^{*}_k,\
  \omega=\sigma(\mathrm{Conv}_{1\times1\times1}([\bar r,\bar r^{*},c])).
\end{equation}
Rotations are fused by quaternion spherical interpolation and step sizes are blended with the same gate:
\begin{equation}
  \bar q=\mathrm{SLERP}(q,q^{*};\omega),\qquad
  \bar{\mathbf{r}}=\omega\mathbf{r}+(1-\omega)\mathbf{r}^{*},\qquad
  \bar R=\mathcal{R}(\bar q).
\end{equation}
The fused field $\bar{\mathcal{H}}=\{\bar R,\bar{\mathbf{r}}\}$ defines the consensus directions and step sizes (via Sec.~\ref{subsec:field})
and is used to steer the stride $1$ refinement on decoder features at skip connection.

\subsection{Implementation notes}\label{subsec:impl}
% Table~\ref{tab:impl_notes} summarises key implementation choices; these do not affect the mathematical definition of the operator in Secs.~\ref{subsec:field}--\ref{subsec:consensus}.
We parameterise $R$ with a unit quaternion (from a $1\times1\times1$ head $H_{\mathrm{geo}}$) and bound $r_k$ to $[r_{\min},r_{\max}]$ via a sigmoid.
Gate is a grouped $1\times1\times1$ conv (one group per token) followed by a $1\times1\times1$ Mix conv; stride $1$ and downsampling use separate heads.
For stability, we stop gradients through $r_k$ in the $1/(r_k+\epsilon)$ factors.

% \begin{table}[t]
% \caption{Key implementation choices.}
% \label{tab:impl_notes}
% \centering
% \fontsize{8}{9}\selectfont
% \resizebox{0.75\textwidth}{!}{%
% \begin{tabular}{ll}
% \hline
% Component & Choice \\
% \hline
% Frame parameterisation & quaternion $\rightarrow$ $R\in SO(3)$ \\
% Step-size bounding & sigmoid to $[r_{\min},r_{\max}]$ \\
% Token set & $\mathcal{T}=[x_0,a_1,\ldots,a_K,G_x,G_y,G_z,L]$, $M=K+5$ \\
% Gate & grouped $1\times1\times1$ conv (groups $=M$) \\
% Mix & $1\times1\times1$ conv \\
% Stride-specific mixers & separate Gate/Mix heads for stride $1$ and $s\neq(1,1,1)$ \\
% Stabilisation & stop-gradient through $r_k$ in $1/(r_k+\epsilon)$ terms \\
% \hline
% \end{tabular}%
% }
% \end{table}

\section{Experiments}
\label{sec:experiments}

\noindent\textbf{Datasets.}
We evaluate our method on three public datasets spanning three different modalities: \textbf{BraTS} (MRI) \cite{baid2021rsna,menze2014multimodal}, \textbf{MSD HepaticVessel} (CT)\cite{antonelli2022medical}, and \textbf{TDSC-ABUS} (Ultrasound) \cite{luo2025tumor}.
HepaticVessel serves as a thin-structure continuity stress test, while BraTS and TDSC-ABUS assess general lesion/tumour segmentation under distinct appearance statistics.
We use a fixed-seed 75/10/15 train/val/test split for all methods.

% \noindent\textbf{Operator-level comparison.}
% Under a fixed 3D U-Net macro-architecture, we compare the baseline, DCNv1/DCNv2 \cite{dai2017deformable,zhu2019deformable}, Dynamic Downsampling (DCD) \cite{yang2025dynamic}, and the proposed operator; our model replaces \emph{all} stride-$1$ blocks and \emph{all} stride-$2$ reductions (Sec. \ref{sec:methodology}). Results are in Table~\ref{tab:op_level}.

% \noindent\textbf{Architecture-level plug-in validation.}
% On TDSC-ABUS we integrate the proposed operator into nnU-Net, Swin-UNETR, and MedNeXt \cite{isensee2021nnu,hatamizadeh2021swin,roy2023mednext}.
% For nnU-Net we replace all stride-$1$ and stride-$2$ blocks with our operator, whereas for Swin-UNETR and MedNeXt we only replace downsampling to preserve their backbone specific refinement design.
% All architectures additionaly apply geometry-consistent alignment after upsampling and before skip connection (Sec.~\ref{subsec:consensus}); results are in Table~\ref{tab:arch_level}.

\noindent\textbf{Training and evaluation.}
All runs use the same training pipeline per dataset with patch-based training and sliding-window inference; patch sizes for Params (model size) / FLOPs (compute cost; $\approx$ number of arithmetic operations) are $128\times192\times128$ (BraTS) and $128^3$ (HepaticVessel / TDSC-ABUS). We report mean Dice and boundary metrics (HD95 and ASSD).

\begin{figure}[t]
  \centering
  \setlength{\tabcolsep}{1.2pt}
  \renewcommand{\arraystretch}{1.0}
  \scriptsize
  \includegraphics[width=0.85\textwidth]{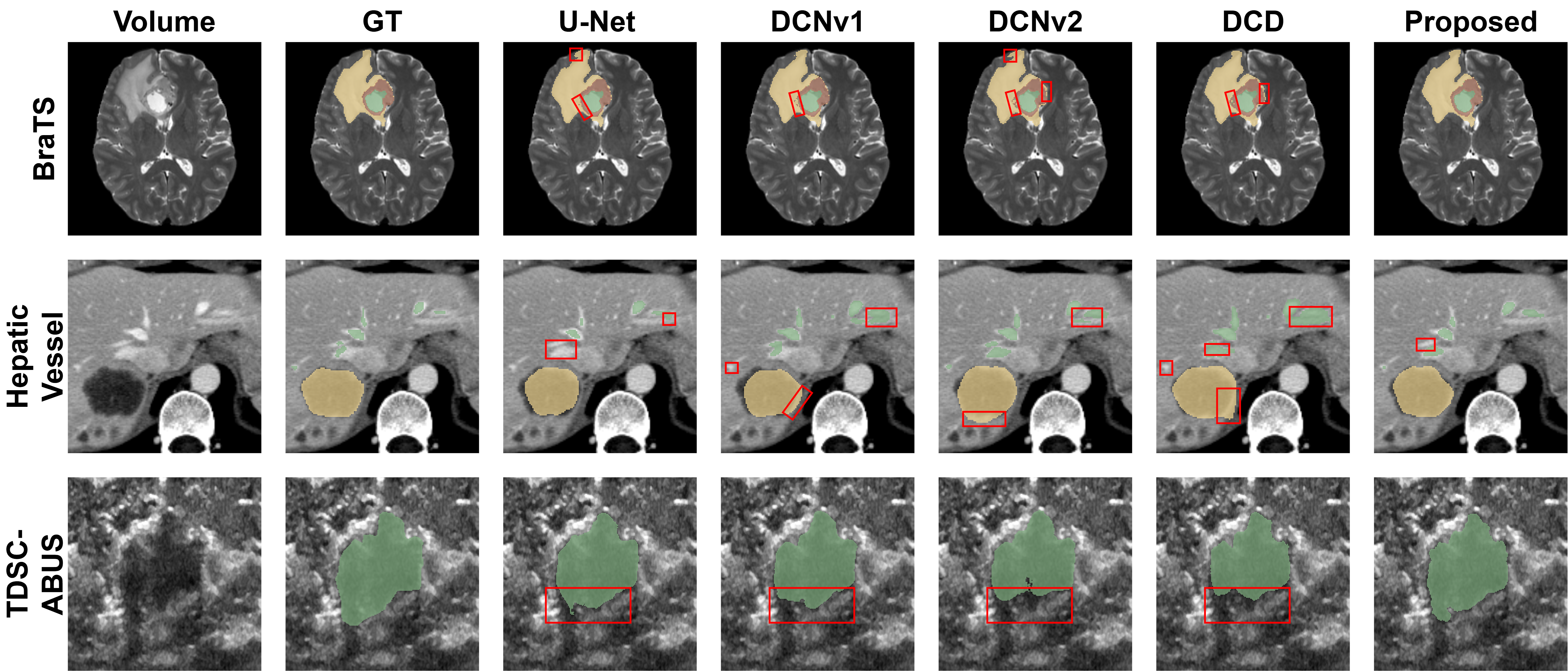}
  \caption{Qualitative comparison on representative cases from BraTS (top), MSD Hepatic Vessel (middle), and TDSC-ABUS (bottom).
  Shown are input, ground truth, and predictions from the baseline U-Net, DCNv1/v2, Dynamic Downsampling (DCD), and the proposed operator; red boxes mark noticeable segmentation errors reduced by our method.}
  \label{fig:qualitative}
\end{figure}

% ===== Operator-level table =====
\begin{table}[!t]
\caption{Operator-level comparison across datasets (top: Dice/HD95/ASSD). Bottom: model complexity (Params/FLOPs) under evaluation patch sizes.}
\label{tab:op_level}
\centering

% ===== Top: metrics =====
\setlength{\tabcolsep}{2.8pt}
\renewcommand{\arraystretch}{1.08}
\resizebox{0.95\textwidth}{!}{%
\small
\begin{tabular}{lcccccccccccc}
\toprule
 & \multicolumn{3}{c}{\shortstack[c]{MSD Hepatic Vessel}}
 & \multicolumn{3}{c}{BraTS}
 & \multicolumn{3}{c}{TDSC-ABUS}
 & \multicolumn{3}{c}{Mean} \\
\cmidrule(lr){2-4}\cmidrule(lr){5-7}\cmidrule(lr){8-10}\cmidrule(lr){11-13}
Model
& \shortstack[c]{Dice$\uparrow$}
& \shortstack[c]{HD95$\downarrow$}
& \shortstack[c]{ASSD$\downarrow$}
& \shortstack[c]{Dice$\uparrow$}
& \shortstack[c]{HD95$\downarrow$}
& \shortstack[c]{ASSD$\downarrow$}
& \shortstack[c]{Dice$\uparrow$}
& \shortstack[c]{HD95$\downarrow$}
& \shortstack[c]{ASSD$\downarrow$}
& \shortstack[c]{Dice$\uparrow$}
& \shortstack[c]{HD95$\downarrow$}
& \shortstack[c]{ASSD$\downarrow$} \\
\midrule
Baseline U-Net
& 53.5 & 42.8 & 8.87
& 86.1 & 7.1 & 1.15
& 64.4 & 39.1 & 7.31
& 68.0 & 29.7 & 5.78 \\
Deformable Conv v1
& \second{57.2} & 37.3 & 7.93
& 88.3 & \second{6.4} & \second{1.07}
& \best{67.5} & \second{33.1} & \second{5.83}
& \second{71.0} & 25.6 & \second{4.94} \\
Deformable Conv v2
& 56.7 & 36.3 & 8.21
& 88.3 & 6.5 & 1.33
& 66.3 & 35.4 & 6.06
& 70.4 & 26.1 & 5.20 \\
\shortstack[l]{Dynamic\\Downsampling}
& 56.4 & \second{34.9} & \best{7.40}
& \second{88.4} & 6.8 & 1.14
& 67.2 & 33.9 & 6.87
& 70.6 & \second{25.2} & 5.13 \\
Proposed
& \best{58.2} & \best{34.3} & \second{7.43}
& \best{88.9} & \best{6.2} & \best{0.95}
& \second{67.4} & \best{27.8} & \best{5.78}
& \best{71.5} & \best{22.8} & \best{4.72} \\
\bottomrule
\end{tabular}%
}

% ===== Bottom: complexity =====
\par\smallskip
\setlength{\tabcolsep}{2.8pt}
\renewcommand{\arraystretch}{1.02}
\resizebox{0.95\textwidth}{!}{%
\small
\begin{tabular}{lcc cc cc cc cc}
\toprule
Input size
& \multicolumn{2}{c}{Baseline U-Net}
& \multicolumn{2}{c}{Deformable Conv v1}
& \multicolumn{2}{c}{Deformable Conv v2}
& \multicolumn{2}{c}{Dynamic Downsampling}
& \multicolumn{2}{c}{Proposed} \\
\cmidrule(lr){2-3}\cmidrule(lr){4-5}\cmidrule(lr){6-7}\cmidrule(lr){8-9}\cmidrule(lr){10-11}
& Params (M)$\downarrow$ & FLOPs (G)$\downarrow$
& Params (M)$\downarrow$ & FLOPs (G)$\downarrow$
& Params (M)$\downarrow$ & FLOPs (G)$\downarrow$
& Params (M)$\downarrow$ & FLOPs (G)$\downarrow$
& Params (M)$\downarrow$ & FLOPs (G)$\downarrow$ \\
\midrule
$128^3$
& 2.3 & 131.2
& 2.5 & 137.9
& 2.6 & 146.9
& 2.7 & 256.7
& 0.8 & 69.9 \\
$128\times192\times128$
& 2.3 & 194.8
& 2.5 & 203.3
& 2.6 & 209.3
& 2.7 & 384.5
& 0.8 & 108.9 \\
\bottomrule
\end{tabular}%
}
\end{table}

\noindent\textbf{Operator-level results.}
Under a fixed 3D U-Net macro-architecture, we compare the baseline convolution, Deformable Convolution (DCN) v1/v2 \cite{dai2017deformable,zhu2019deformable}, Dynamic Downsampling (DCD) \cite{yang2025dynamic}, and the proposed operator by replacing all stride $1$ blocks and stride $2$ reductions with the respective operator (Table~\ref{tab:op_level}).
For BraTS, metrics are averaged over enhanced tumour, tumour core and whole tumour; for Hepatic Vessel, we calculated mean metrics on vessel and tumour; for TDSC-ABUS, we report metrics on tumour.
Our operator achieves the best mean Dice/HD95/ASSD ($71.5/22.8/4.72$).
It is best on BraTS ($88.9/6.2/0.95$), attains the best Dice/HD95 on Hepatic Vessel ($58.2/34.3$) with near-best ASSD ($7.43$ vs.\ $7.40$), and markedly improves boundary metrics on TDSC-ABUS (HD95/ASSD $27.8/5.78$) while keeping Dice comparable ($67.4$), indicating the most stable cross-modality improvements overall.

\noindent\textbf{Computational Complexity.}
Table~\ref{tab:op_level} (bottom) reports Params and FLOPs in the U-Net setting.
Our operator reduces Params from $2.3$M to $0.8$M and FLOPs from $194.8$G to $108.9$G.
By contrast, DCNv1/v2 slightly increases both, while Dynamic Downsampling raises FLOPs to $256.7$G/$384.5$G.
This efficiency comes from using a small number of symmetric samples (\eg, $K\!=\!3$ paired directions) and compact token mixing with $1\!\times\!1\!\times\!1$ gating, rather than dense grids or heavy dynamic branches.

% =========================
% Architecture-level table
% =========================
\begin{table}[!t]
\caption{Architecture-level plug-in validation on TDSC-ABUS dataset.}
\label{tab:arch_level}
  \centering
  \setlength{\tabcolsep}{4.5pt}
  \resizebox{0.95\textwidth}{!}{%
  \small
    \begin{tabular}{ll
                    S[table-format=1.3]
                    S[table-format=2.2]
                    S[table-format=2.2]
                    S[table-format=2.2]
                    S[table-format=2.2, round-precision=2]
                    S[table-format=2.2]
                    S[table-format=4.0]}
      \toprule
      Backbone & Variant &
      \multicolumn{3}{c}{Overlap} &
      \multicolumn{2}{c}{Boundary} &
      \multicolumn{2}{c}{Efficiency} \\
      \cmidrule(lr){3-5}\cmidrule(lr){6-7}\cmidrule(lr){8-9}
      & &
      {Dice$\uparrow$} &
      {Sensitivity$\uparrow$} &
      {Specificity$\uparrow$} &
      {HD95$\downarrow$} &
      {ASSD$\downarrow$} &
      {Params (M)$\downarrow$} &
      {FLOPs (G)$\downarrow$} \\
      \midrule
      \multirow{2}{*}{nnU-Net}
        & Baseline           & 68.0 & 72.4 & 98.8 & 28.4 & 5.24 & 30.8 & 1250 \\
        & + Proposed         & \best{72.8} & 75.3 & 98.5 & \best{25.6} & \best{5.03} & 17.4 & 750.2 \\
      \midrule
      \multirow{2}{*}{Swin-UNETR}
        & Baseline           & 69.1 & 68.0 & 98.6 & 30.2 & 4.97 & 63.5 & 751.5 \\
        & + Proposed         & \best{70.9} & 74.2 & 98.7 & \best{27.7} & \best{4.88} & 64.8 & 802.7 \\
      \midrule
      \multirow{2}{*}{MedNeXt}
        & Baseline           & 66.8 & 69.5 & 99.1 & 29.4 & 5.38 & 2.71 & 50.1 \\
        & + Proposed         & \best{69.5} & 75.5 & 98.2 & \best{24.2} & \best{4.89} & 2.89 & 57.0 \\
      \bottomrule
    \end{tabular}%
  }
\end{table}

\begin{figure}[t]
  \centering
  \setlength{\tabcolsep}{1.2pt}
  \renewcommand{\arraystretch}{1.0}
  \scriptsize
  \includegraphics[width=0.85\textwidth]{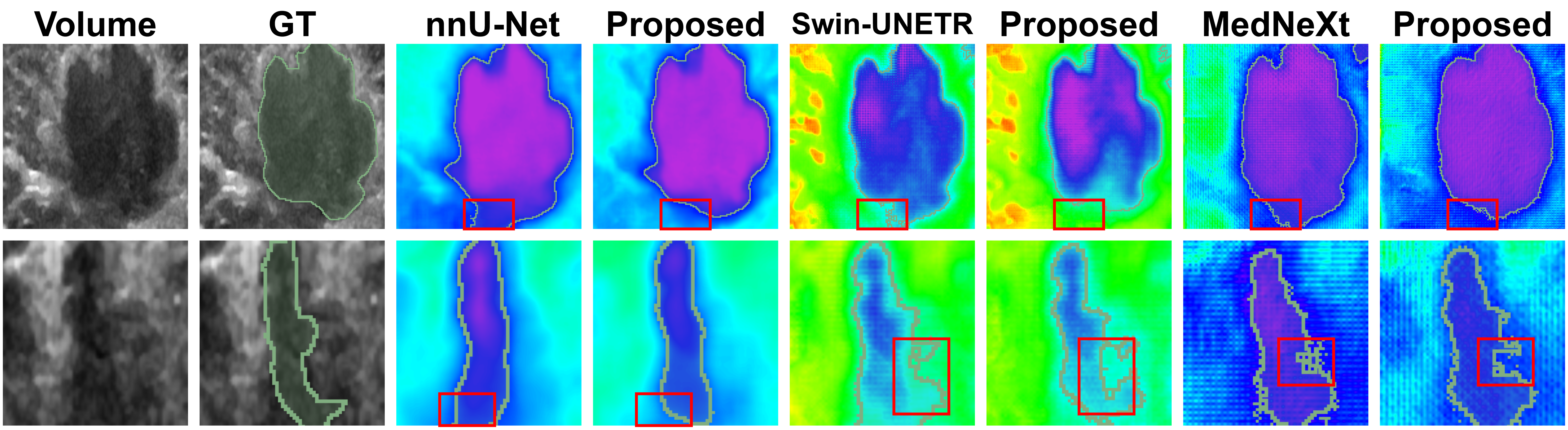}
  \caption{Qualitative plug-in results on TDSC-ABUS (red boxes: improved regions).}
  \label{fig:qualitative2}
\end{figure}

\begin{figure}[t]
  \centering
  \setlength{\tabcolsep}{1.2pt}
  \renewcommand{\arraystretch}{1.0}
  \scriptsize
  \includegraphics[width=0.85\textwidth]{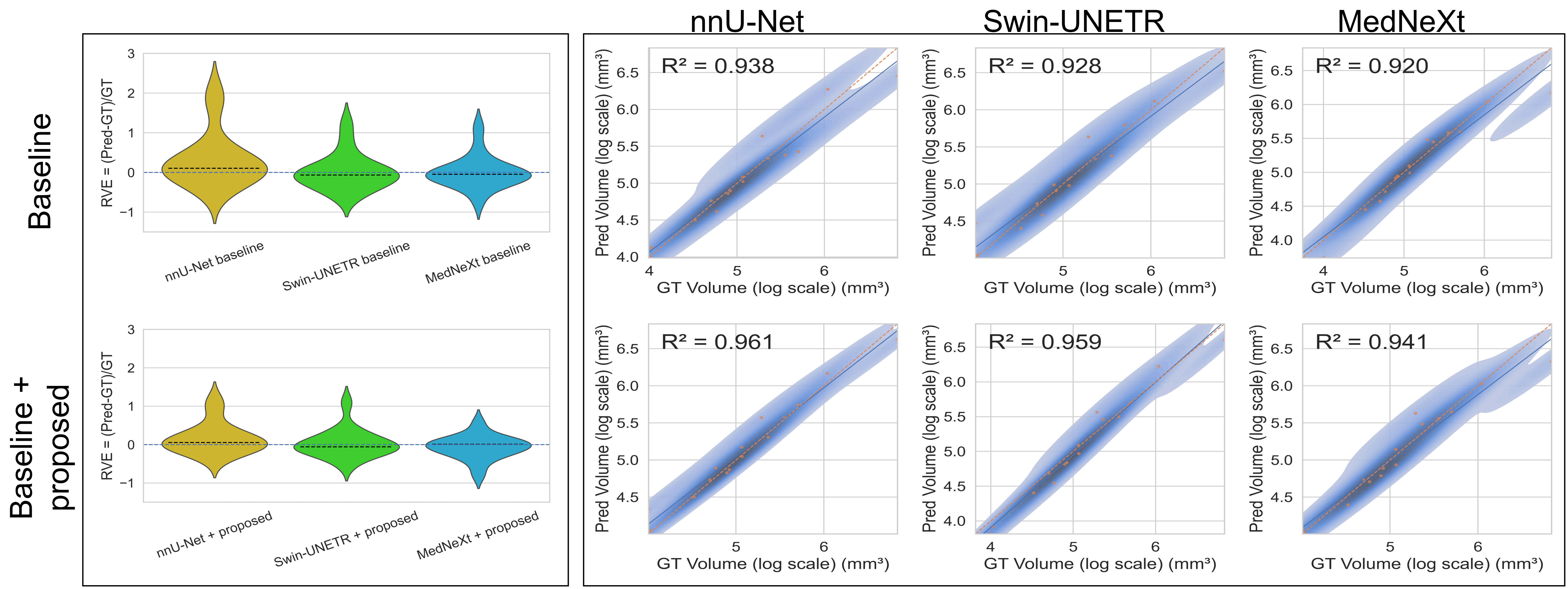}
  \caption{Volumetric agreement on TDSC-ABUS. Left: relative volume error (RVE) distributions for different backbones. Right: predicted vs.\ ground-truth tumour volumes (log scale) for baseline and +Proposed; dashed line denotes $y=x$.}
  \label{fig:qualitative3}
\end{figure}

% \begin{figure}[t]
%   \centering
%   \setlength{\tabcolsep}{1.2pt}
%   \renewcommand{\arraystretch}{1.0}
%   \scriptsize

%   \begin{minipage}{0.75\textwidth}
%     \centering
%     \includesvg[width=\linewidth]{figures/qualitative_figure3}
%   \end{minipage}

%   \begin{minipage}{0.75\textwidth}
%     \centering
%     \includesvg[width=\linewidth]{figures/volume}
%   \end{minipage}

%   \caption{Plug-in results on TDSC-ABUS. Top: qualitative comparison (red boxes: reduced errors). Bottom: predicted vs.\ ground-truth tumour volumes (log scale; dashed: $y=x$).}
%   \label{fig:plugin_tdsc}
% \end{figure}

\noindent\textbf{Architecture-level plug-in results.}
On TDSC-ABUS we plug the proposed operator into nnU-Net, Swin-UNETR, and MedNeXt (base version) \cite{isensee2021nnu,hatamizadeh2021swin,roy2023mednext} (Table~\ref{tab:arch_level}).
For nnU-Net we replace all stride $1$ and stride $>1$ blocks, whereas for Swin-UNETR and MedNeXt we only replace downsampling to preserve backbone-specific refinement; all variants additionally apply \emph{Consensus Field} at skip connection (Sec.~\ref{subsec:consensus}).
Across backbones, the plug-in improves overlap and boundary metrics (Dice $+1.8$--$+4.8$, Sens. $+2.9$--$+6.2$; HD95 $-2.5$--$-5.2$, ASSD $-0.09$--$-0.49$), with minor specificity changes ($\ge$98.2\%).
The largest gain is on nnU-Net (Dice $68.0\!\rightarrow\!72.8$) while reducing cost (Params $30.8$M$\!\rightarrow\!17.4$M; FLOPs $1250$G$\!\rightarrow\!750$G); Swin-UNETR and MedNeXt also improve with modest FLOPs changes, consistent with their already-efficient downsampling (patch merging / depthwise separable convolutions).
Fig.~\ref{fig:qualitative2} shows fewer leakage/boundary errors, and Fig.~\ref{fig:qualitative3} indicates improved volumetry (RVE distributions closer to 0 with fewer outliers and points closer to $y=x$).
Overall, the improved boundaries and volumetry suggest more reliable downstream quantitative assessment across backbones.

\noindent\textbf{Ablation Study.}
We test the effectiveness of the geometry-aware modules. Under the same 3D U-Net setting using TDSC-ABUS data, removing the differentials ($G,L$; geometry-aware gradient/curvature cues) drops Dice/HD95/ASSD from $67.4/27.8/5.78$ to $54.3/51.0/10.71$, and removing the \emph{Consensus Field} (skip alignment to reduce geometry mismatch) drops to $58.4/48.2/8.52$. 
This confirms that geometric differential terms drive boundary quality, with consensus alignment providing additional cross-scale robustness.

\section{Conclusion}\label{sec:conclusion}
We have introduced \emph{GeoSample}, a geometry-guided operator for volumetric encoder--decoder segmentation that predicts local orientations to steer symmetric sampling and extract geometric signals for refinement and downsampling.
Across three public benchmarks and plug-in validation, it consistently improves segmentation quality, especially boundary metrics, while reducing the number of parameters in the U-Net setting.

Future work will focus on reducing the memory/wall-clock time overhead of interpolation-based sampling and improve early-training stability.
\bibliographystyle{splncs04}
\bibliography{references}
%
% \begin{thebibliography}{8}
% \bibitem{ref_article1}
% Author, F.: Article title. Journal \textbf{2}(5), 99--110 (2016)

% \bibitem{ref_lncs1}
% Author, F., Author, S.: Title of a proceedings paper. In: Editor,
% F., Editor, S. (eds.) CONFERENCE 2016, LNCS, vol. 9999, pp. 1--13.
% Springer, Heidelberg (2016). \doi{10.10007/1234567890}

% \bibitem{ref_book1}
% Author, F., Author, S., Author, T.: Book title. 2nd edn. Publisher,
% Location (1999)

% \bibitem{ref_proc1}
% Author, A.-B.: Contribution title. In: 9th International Proceedings
% on Proceedings, pp. 1--2. Publisher, Location (2010)

% \bibitem{ref_url1}
% LNCS Homepage, \url{http://www.springer.com/lncs}, last accessed 2023/10/25
% \end{thebibliography}
\end{document}